\documentclass{article}
\usepackage{spconf,amsmath,graphicx}
\usepackage{amssymb}
\usepackage{diagbox}
\usepackage{blindtext}

\usepackage{soul}
\usepackage{color}

\title{SINGLE IMAGE CALIBRATION USING KNOWLEDGE DISTILLATION APPROACHES}
\name{Khadidja Ould Amer$^{1}$, Oussama Hadjerci$^{1}$, Mohamed Abbas Hedjazi$^{1}$, Antoine Letienne$^{2}$}
\address{$^{1}$DASIA, Corbreuse, France.\\
$^{2}$ CLIRISGROUP, Courbevoie, France. }

\begin{document}
%
\maketitle
\begin{abstract}
Although recent deep learning-based calibration methods can predict extrinsic and intrinsic camera parameters from a single image, their generalization remains limited by the number and distribution of training data samples. The huge computational and space requirement prevents convolutional neural networks (CNNs) from being implemented in resource-constrained environments. This challenge motivated us to learn a CNN gradually, by training new data while maintaining performance on previously learned data. 
Our approach builds upon a CNN architecture to automatically estimate camera parameters ($focal$ $length$, $pitch$, and $roll$) using different incremental learning strategies to preserve knowledge when updating the network for new data distributions. Precisely, we adapt four common incremental learning, namely: $LwF$, $iCaRL$, $LUCIR$, and $BiC$ by modifying their loss functions to our regression problem. We evaluate on two datasets containing $299008$ indoor and outdoor images. Experiment results were significant and indicated which method was better for the camera calibration estimation.

\end{abstract}

\section{Introduction}
\label{sec:intro}

Single image calibration estimates camera parameters from a monocular RGB image. This problem is of significant importance in many computer vision tasks, especially in applications where capturing conditions are not controlled \cite{fischer2015image,karsch2014automatic}. Conventional approaches of single image camera calibration rely on detecting reference objects in the scene, such as a calibration grid \cite{mei2007single} or co-planar circles \cite{chen2004camera}. Other methods take advantage of vanishing point properties, by carefully selecting parallel or orthogonal segments in the 3D scene \cite{zhu2020single}. However, most of these methods use classic image processing techniques to detect geometric cues, which makes them inapplicable in unstructured environments. To overcome such limitations, recent studies exploited semantic cues learned by deep neural networks to estimate camera parameters. Existing approaches focus on estimating intrinsic \cite{workman2015deepfocal,bogdan2018deepcalib} and/or extrinsic parameters \cite{workman2016horizon}. Yannick et al. \cite{hold2018perceptual} estimate the horizon line as a proxy of camera parameters. This joint estimation results significantly improve the performance of the model. The latter is partially due to the generated large-scale synthetic dataset from panoramic images. Despite the success of these methods, their generalization requires a very large-scale dataset covering a large range of image distributions, such as scene types, visible objects, camera types, etc. In practice, the calibration network must be updated to the application-specific data stream to provide more accurate estimations of camera parameters. However, legacy data may be unrecorded, proprietary, or simply too cumbersome to use in training a new task. As a consequence, it becomes apparent that more flexible strategies are required to handle the large-scale and dynamic properties of real-world situations. 

In this work, we adopt a CNN architecture to recover extrinsic ($pitch$, $roll$) and intrinsic ($focal$ $length$) camera parameters from a single input image. Our main contribution is to explore methods to improve the generalization capability of our network. Specifically, we consider two data distribution sets (indoor and outdoor scene). We train our network exclusively on a single dataset and use different incremental learning strategies to preserve acquired knowledge when training on the second dataset. To our best knowledge,  this is the first study that applies incremental learning techniques for the camera parameters estimation.

\section{Camera model}
\label{sec:Related Work}
The process of image formation has been studied extensively in computer vision \cite{andrew2001multiple}, allowing for very precise calibration of the camera when there are
enough constraints to fit the  geometric camera model. Under the pinhole camera model, homogeneous pixel coordinates $p_{im}$ of a $3D$ point $p_w$ are given by:
\begin{equation}
p_{im}=[\lambda_{u},\lambda_{v},\lambda]=[R \mid t][p_w \mid 1]^T
\end{equation}
where $K$ is the camera projection matrix (camera intrinsics), $R$ and $t$ are the camera rotation and translation in the world reference system (camera extrinsics). By considering square pixels, no skew, and image center at the principal point, the
projection matrix $K$ can be symplified to $K = diag([f_{px}, f_{px}, 1])$, where $f_{px}$ is the focal length in pixels.
The rotation matrix $R$, can be parameterized by roll $\psi$, pitch $\theta$, and yaw $\varphi$ angles. There exists no natural reference frame to estimate $\varphi$ (left vs right) from an arbitrary image. Therefore, the extrinsic rotation matrix is constrained to only pitch and roll components $(R = R_z (\psi)R_x (\theta))$.

For the case of camera calibration, parameters such as focal length, pitch and roll are difficult to interpret from image content. As revealed by previous works \cite{workman2016horizon,hold2018perceptual}, we can use horizon line as an intuitive representation for these
parameters. We define the horizon line midpoint $b_p$ as the y-coordinate of its intersection with the central vertical axis in the image. It can be derived from $\theta$ and $f _{px}$ as :

\begin{equation}
b_p=2f_{p_x}tan(\theta)
\label{bp}
\end{equation}
The roll angle $\psi$ represents the angle between the horizon line and the horizontal axis of the image. In this image unit representation, the top and bottom of the image have coordinates $1$ and $-1$ respectively.

\section{Image calibration network}
\label{ICN}
Our goal is to train a CNN to estimate camera parameters from a single image. To achieve this, we use independent regressors that share a common pretrained network architecture, which we update using our data. Precisely, we adopt Residual Neural Networks with $50$ layers
(ResNet50) \cite{he2016deep}, on which the last layer is replaced with three outputs estimating $focal$, $pitch$ and $roll$ values. In this work, we specifically aim to learn a generalized model, which can efficiently estimate camera parameters whether for indoor or outdoor images. To conduct this evaluation, we train our network using exclusively indoor or outdoor data and exploit a previously trained network by transferring the learned features to the target network. However, fine-tuning the weights of a pretrained network usually degrades the performance on the previously learned task because the shared parameters change without any guidance for the original task. To avoid this problem, we use incremental learning techniques to adapt a CNN to different distributions.

\subsection{Incremental learning}
\label{Incremental_learning}
In the real world, many computer vision applications require learning new visual capabilities while maintaining performance on existing ones. However, CNN-based systems often suffer from the \textit{“catastrophic forgetting”} of the previous knowledge. This problem is mainly due to two facts: (1) the updates can override the knowledge acquired from the previous data, and (2) the model can not replay the entire previous data to regain the old knowledge. To overcome these limits, several methods of incremental learning have been proposed, which are generally divided into three categories: (1) regularization-based methods: use regularization terms in the loss function to alleviate forgetting \cite{li2017learning,hou2019learning,javed2018revisiting,Wu_2019_CVPR,douillard2020podnet}; (2) rehearsal-based methods: construct a small exemplar set from old data \cite{rebuffi2017icarl,hou2019learning} or synthesize samples to keep the performance for old classes \cite{shin2017continual,he2018exemplar}; (3) dynamic architecture methods: utilize different network parameters for different tasks \cite{yoon2017lifelong,li2019learn}.

In this work, we consider a single incremental learning task: we have a regression model already trained on old data, and we need to update it to new data distribution. To solve this problem, we use methods that introduce regularization terms in their loss functions to consolidate previous knowledge when learning from new data. Following, we introduce the common loss distillation-based approaches and their adaptation to our regression problem.

\subsubsection{Knowledge Distillation}
\label{sec:Knowledge Distillation}
Knowledge distillation is first proposed in \cite{hinton2015distilling} to transfer knowledge from a large pre-trained teacher network (or network assembly) to a smaller student network for more efficient deployment. Thereafter, knowledge distillation was introduced in $LwF$ (learning without forgetting) to avoid catastrophic forgetting by appointing a previous snapshot of the model as a teacher while new tasks are learned \cite{li2017learning}. More precisely, $LwF$ preserves the outputs of the old model by optimizing a loss function defined as:
\begin{equation}
\mathcal{L}_{KD}=\mathcal{L}_{new}(y_{n},\hat{y}_{n})+ \lambda_0 \mathcal{L}_{old}(y_{o},\hat{y}_{o})
\label{KD_eq}
\end{equation}

\noindent where: $\mathcal{L}_{new}$ is the common cross-entropy, which encourages new task predictions $\hat{y}_{n}$ to be consistent with ground truths $y_{n}$; $\mathcal{L}_{old}$ is a distillation loss used to prevent the actual outputs of the updated network $\hat{y}_{o}$ from deviating too much from stored outputs of its older version; ${y}_{o}$ and $\lambda_0$ is a loss balance weight. The original distillation loss function is based on a modified cross-entropy loss that produces a softer probability distribution over classes \cite{hinton2015distilling}. In our case, we solve our regression problem using a smooth-L1 loss (Eq. \ref{smooth}), which is less sensitive to outliers than the \textit{mean square error} loss. 
\begin{equation}
\mathcal{L}(y_,\hat{y})= \frac{1}{n}\sum_{i}^{L}
\left \{
\begin{array}{l l}
0.5(y-\hat{y})^{2}  \quad \text{if} \quad &  \mid y-\hat{y} \mid < 1 \\
\mid y-\hat{y} \mid - 0.5 &  \text{otherwise,} 
\end{array}
\right.
\label{smooth}
\end{equation}
Where $n$ is the number of regression outputs, and $L$ is the batch size.

\subsubsection{Replay-based methods}
\label{Replay_based}
$LwF$ is the first work addressing catastrophic forgetting in class incremental learning. Recent studies show that selecting a few exemplars from the old classes can alleviate the imbalance problem. Incremental classifier and representation learning $iCaRl$ \cite{rebuffi2017icarl} employs the nearest mean of exemplars classification strategy to select the most representative samples of each previous class and replay them together with the learning data of new tasks. Learning a unified classifier incrementally via rebalancing $Lucir$ \cite{hou2019learning} uses the $iCaRL$ baseline to select exemplars and incorporates a series of regularization terms to mitigate negative effects caused by data imbalance. Precisely, they apply a cosine normalization in the last layer of the network to make the magnitudes of old and new data predictions more comparable and introduce a less forget constraint to enforce the previous knowledge. They also incorporate a margin ranking loss to avoid ambiguities between old and new data.

In this work, we adapt $Lucir$ to resolve our regression problem by (1) applying cosine normalization in the last layer of ResNet50; (2) incorporating the less forget constraint into our loss function. 
Formally, we calculate $\mathcal{L}_{lucir}$ by summing $\mathcal{L}(y,\hat{y})$ and $\mathcal{L}_{dist}$, where $\mathcal{L}_{dist}= 1-<(\bar{f^*}(x),\bar{f(x)})>$. \noindent $\bar{f^*}(x)$ and $\bar{f(x)}$ are respectively, the normalized features extracted by the original model and those by the current one. 
\begin{table*}[h]
	\caption{$MSE$ errors of calibration parameters obtained using indoor and outdoor model.}
	\centering
	\begin{tabular}{|c|c|c|c|c||c|c|c|c|}
		\hline
		\diagbox{
			Evaluation}{} &
		\multicolumn{4}{c||}{Indoor model} & \multicolumn{4}{c|}{Outdoor model} \\
		\cline{2-9}
		
		& \small{Focal}& \small{Roll} & \small{Pitch} & $\mu MSE$  & \small{Focal}& \small{Roll} & \small{Pitch} & $\mu MSE$ \\
		\hline
		
		Validation indoor &0.1&1.33&0.07&\textbf{0.50}&1.59&2.43&0.75&1.59\\
		\hline
		Validation outdoor &2.17&2.84&0.58&1.86&0.08&1.40&0.08&\textbf{0.52}\\
		\hline
		\hline
		Test indoor &0.09&0.39&0.06&\textbf{0.18}&1.1&2.88&0.91&1.63\\
		\hline
		Test outdoor &2.17&2.84&0.58&1.86&0.09&1.46&0.06&\textbf{0.54}\\
		\hline
	\end{tabular}
	\label{tab_cross}
\end{table*}
\subsubsection{Bias Correction (BiC)}
Most data replay-based incremental learning methods follow the $iCaRL$ benchmark protocol to select exemplars. Recent studies show that  approaches selecting few exemplars from old classes perform well on small datasets. However, they suffer from significant performance degradation when the number of classes becomes large. To solve this issue, a bias correction model called BiC was introduced in \cite{wu2019large}. 
The training process consists of two stages: (1) learning the convolution layers of the network, and (2) applying a linear model to the last FC layer to correct the bias in the validation data, which approximates the real distribution of both old and new classes. 

In this work, we adapt our network to perform bias correction by applying a BiC linear model ($q_k= \alpha o_k+\beta$) to the last FC layer of our network, where $\alpha$ and $\beta$ are the BiC model parameters and  $o_k$ is the k-th regression output, where $k=3$ for focal, pitch and roll. In the training phase, we use the distillation function (equation. \eqref{KD_eq}) for the first stage. For the second stage, we freeze convolution and FC layers and optimize $\mathcal{L}(q,y)$ loss function \eqref{smooth}). 

\section{Experimentation}
In this section, we begin by explaining the generation of the datasets, then we evaluate and compare the methods introduced in section \ref{ICN}. Our experiments are divided into two main parts: (1) cross-evaluation of indoor and outdoor models; (2) network generalization. 

\subsection{Dataset generation}
\label{sec:Dataset  generation}
\begin{figure}[h!]
	\centering
	\includegraphics[height= 0.245\columnwidth]{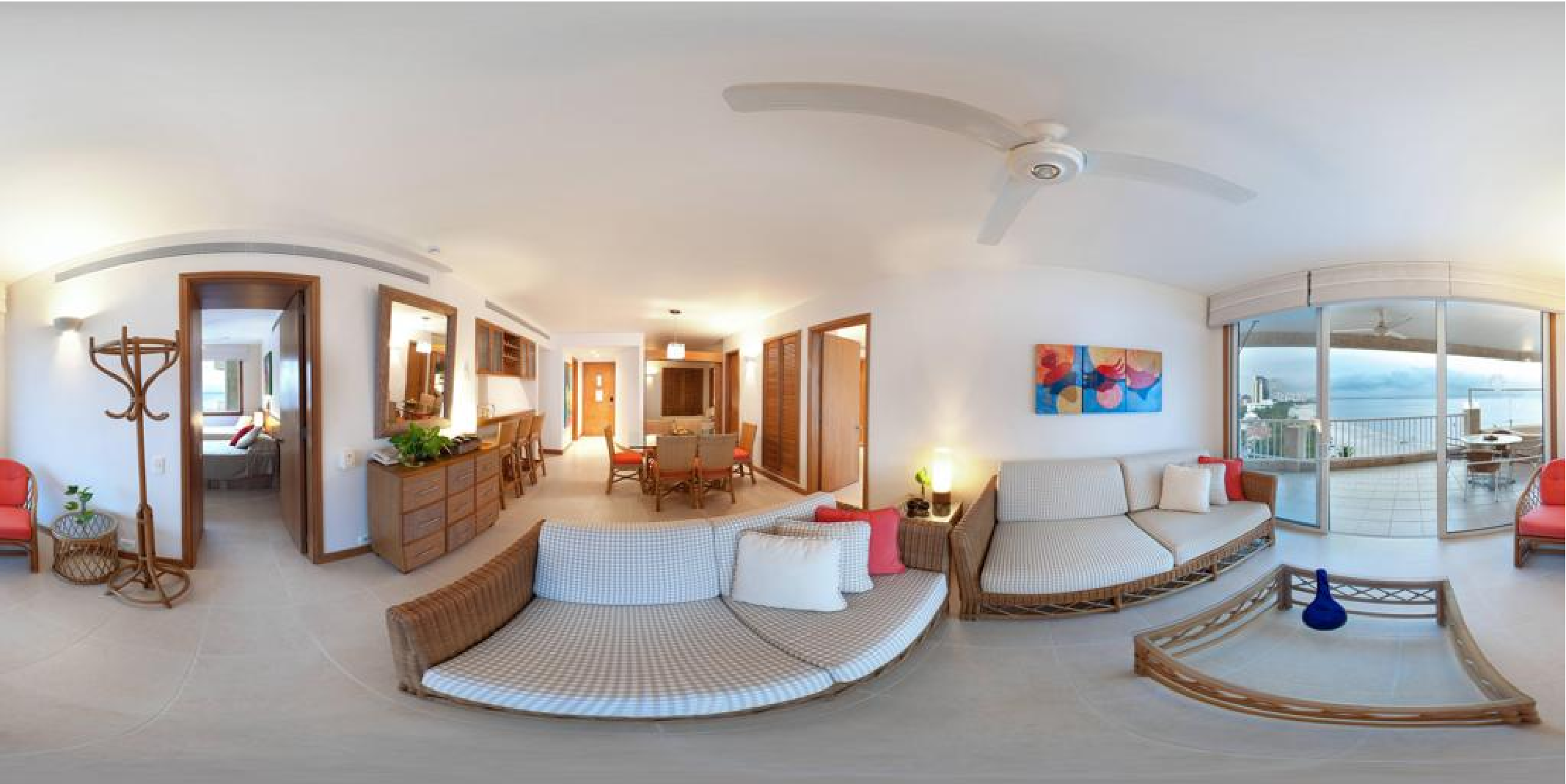}
	\includegraphics[height= 0.245\columnwidth]{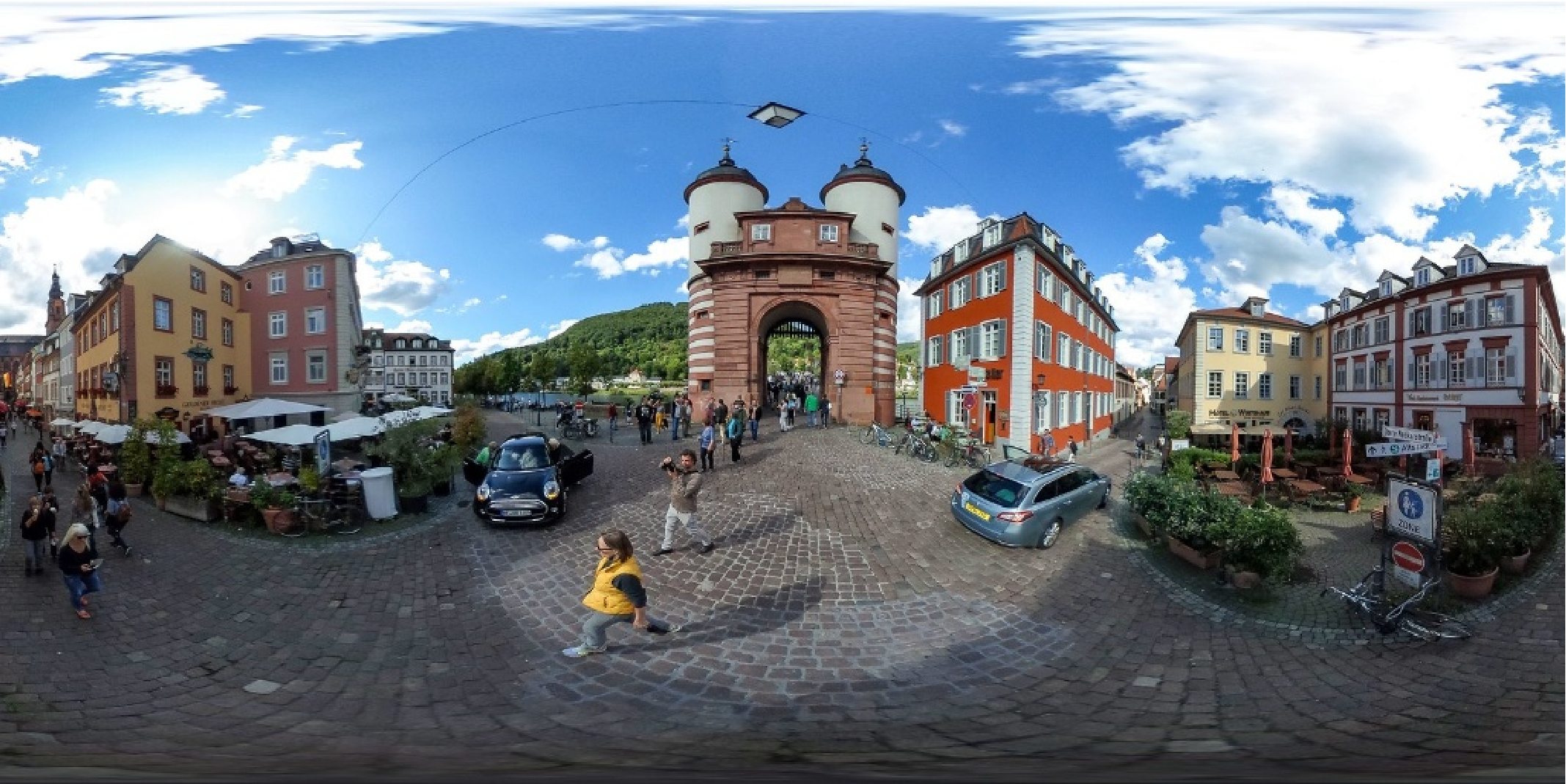}
	\caption{Examples of panoramas used to generate synthetic training datasets : (left) indoor image, (right) outdoor image.} 
	\label{fig:in_out}
\end{figure}

We took inspiration from \cite{hold2018perceptual} to synthetically generate images and their ground truth camera parameters. In this work, we divided the data into two categories: indoor and outdoor datasets (Fig.\ref{fig:in_out}). 
To construct the indoor dataset, we choose $292$ panoramic images from the publicly available SUN360 database \cite{xiao2012recognizing}. Furthermore, we collect $292$ public panoramic images from the internet to construct the outdoor dataset (we particularly chose images of the same resolution as the indoor SUN360 images $[1024 \times 512px]$). On one hand, panoramic images are used to emulate any amount of $360 ^\circ$ field of view. On the other hand, we can point the virtual camera to different orientations to observe different parts of the scene and mimic tilted cameras. 

In this work, we randomly generate focal lengths in a range ($50$,$500$ pixel). For rotation angles, 
we randomly generate pitch $\theta$ in range $(-90^\circ,0^\circ)$ with a roll angle $\psi$ comprised between $(-45^\circ,45^\circ)$. We divided ground truths by the max absolute value of the corresponding parameter to obtain a comparable scales: $focal \in [0,1], pitch \in [-1, 0], roll \in [-1,1]$.
By following this approach, we automatically generated two databases (indoor and outdoor) each containing $149504$ images, 80\% were used for training and 20\% for validation. Similarly, we generate two test sets with $3584$ images each, where the size of images is $299$ $\times$ $299$ pixels. 

\subsection{Cross-evaluation}
We conduct our experiments by a cross-evaluation of indoor and outdoor models (as stated in section \ref{ICN}) using exclusively indoor images and we test the resulting network on indoor and outdoor datasets. Conversely, we train our network on outdoor images and test the resulting model on the two datasets. ResNet50 is trained to minimize the L1-smooth loss between ground truths and predictions. The model is trained on a single GPU with a batch size of $16$. We use the SGD optimizer with an initial learning rate of $0.003$, which is reduced by a factor of $0.1$ if no improvement is seen over $2$ epochs. 

The results of cross-testing are reported in Table.\ref{tab_cross}. We specifically show the \textit{mean squared error} ($MSE$) of the camera parameters ($focal$, $pitch$, $roll$) and compare results according to the mean $MSE$ of the three parameters that we call $\mu MSE$. 
\begin{figure}[h!]
	\centering
	\includegraphics[height= 0.32\columnwidth]{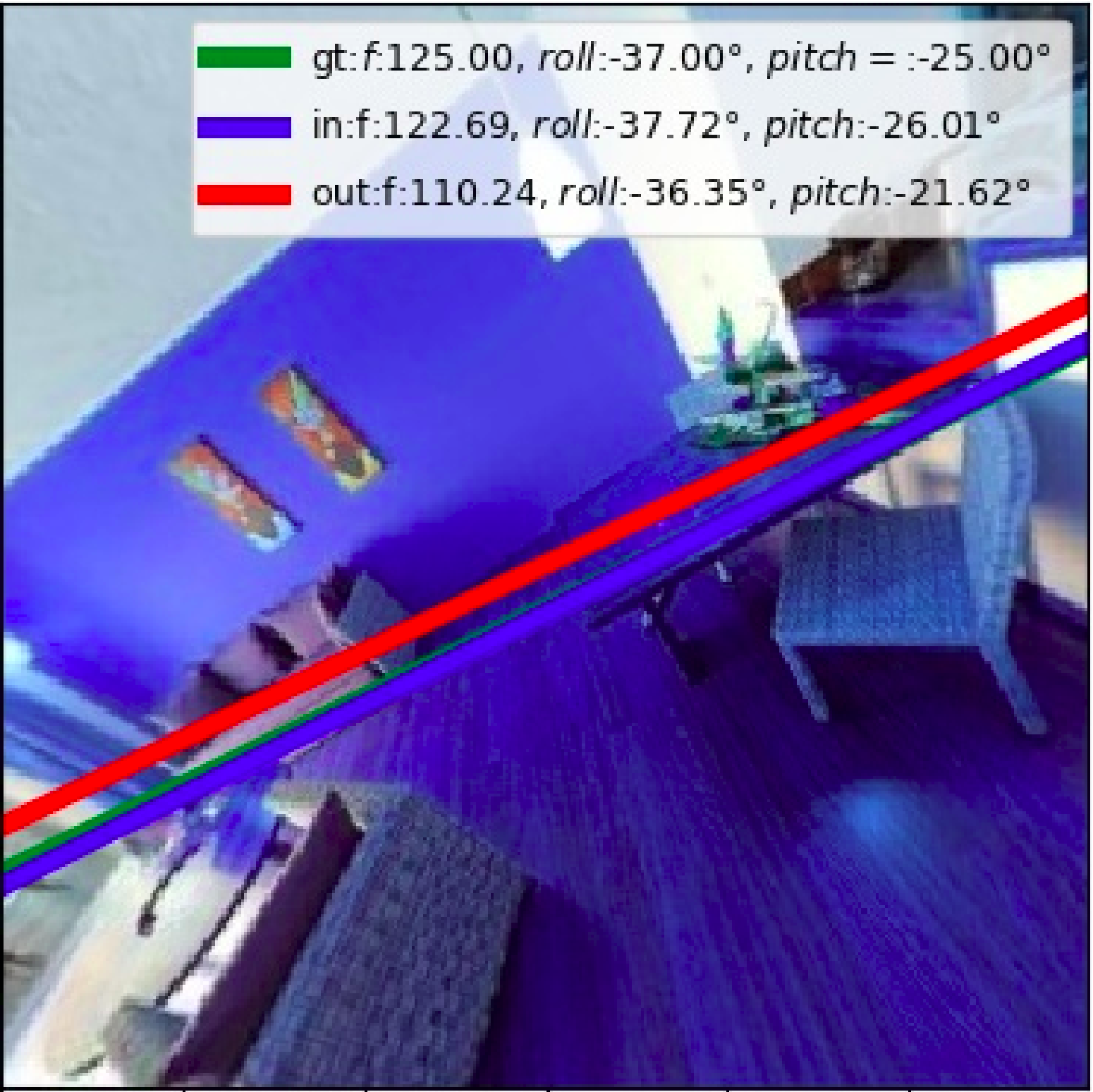}
	\includegraphics[height= 0.32\columnwidth]{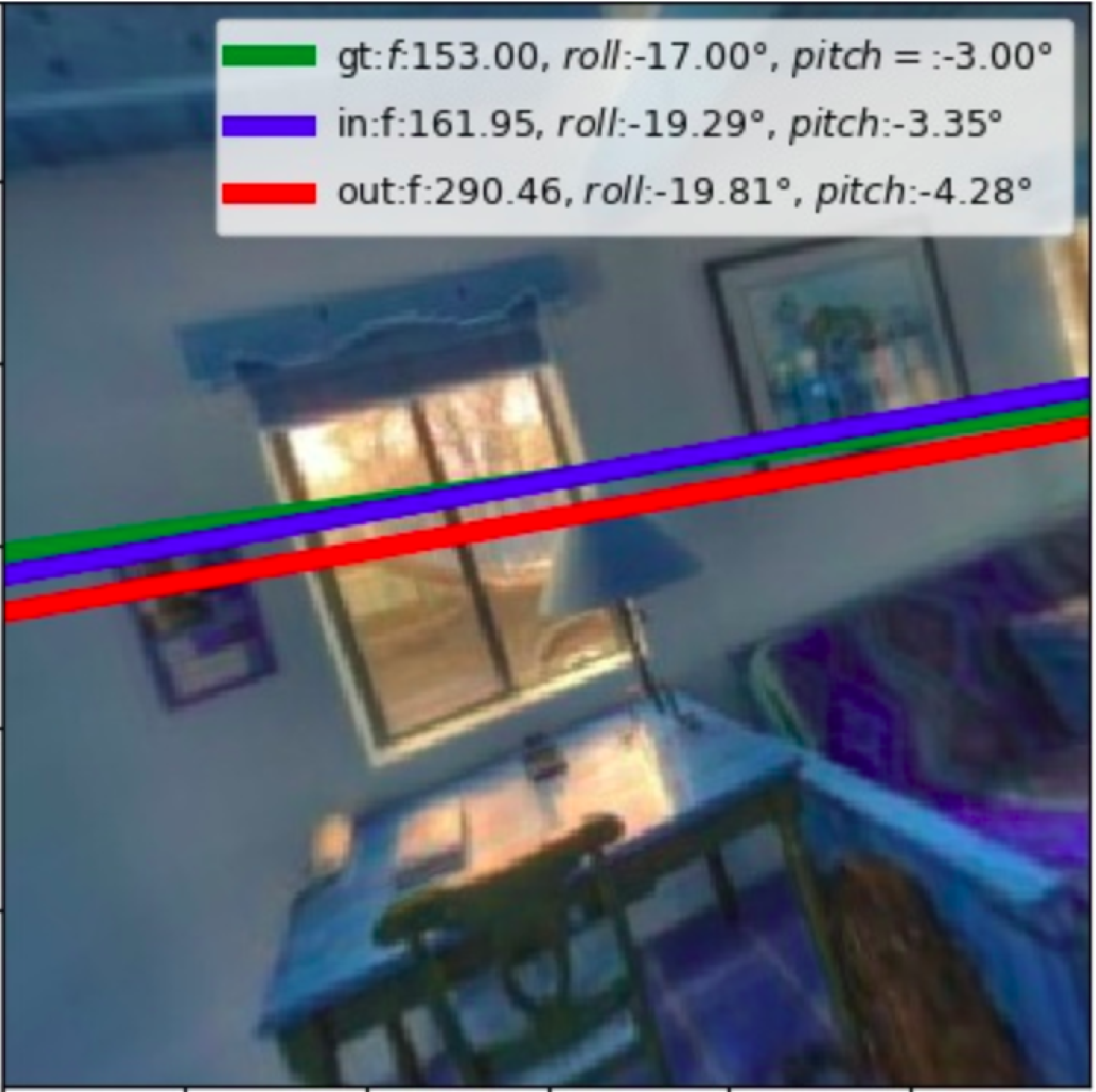}
	\includegraphics[height= 0.32\columnwidth]{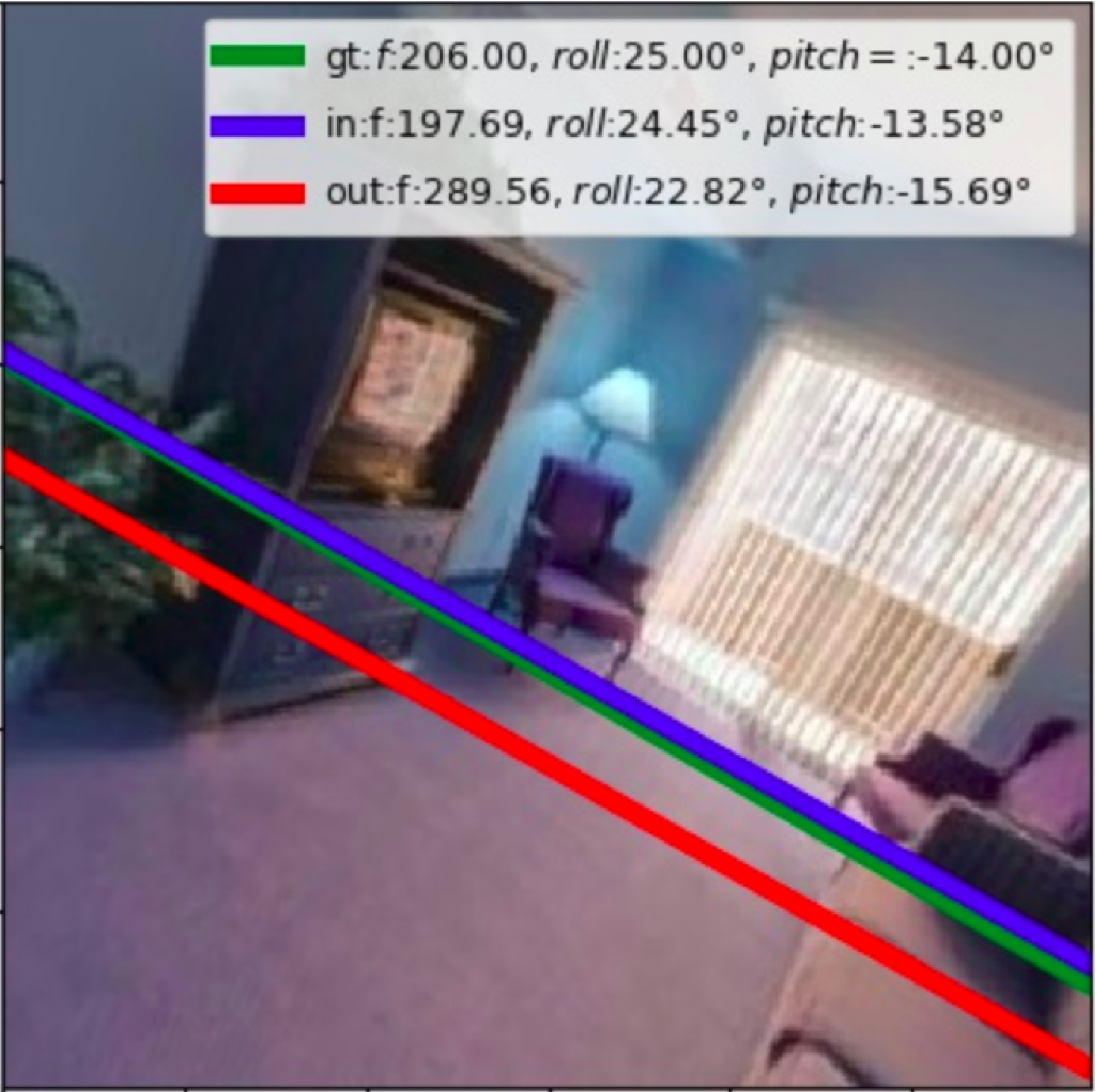}
	\\
	\includegraphics[height= 0.32\columnwidth]{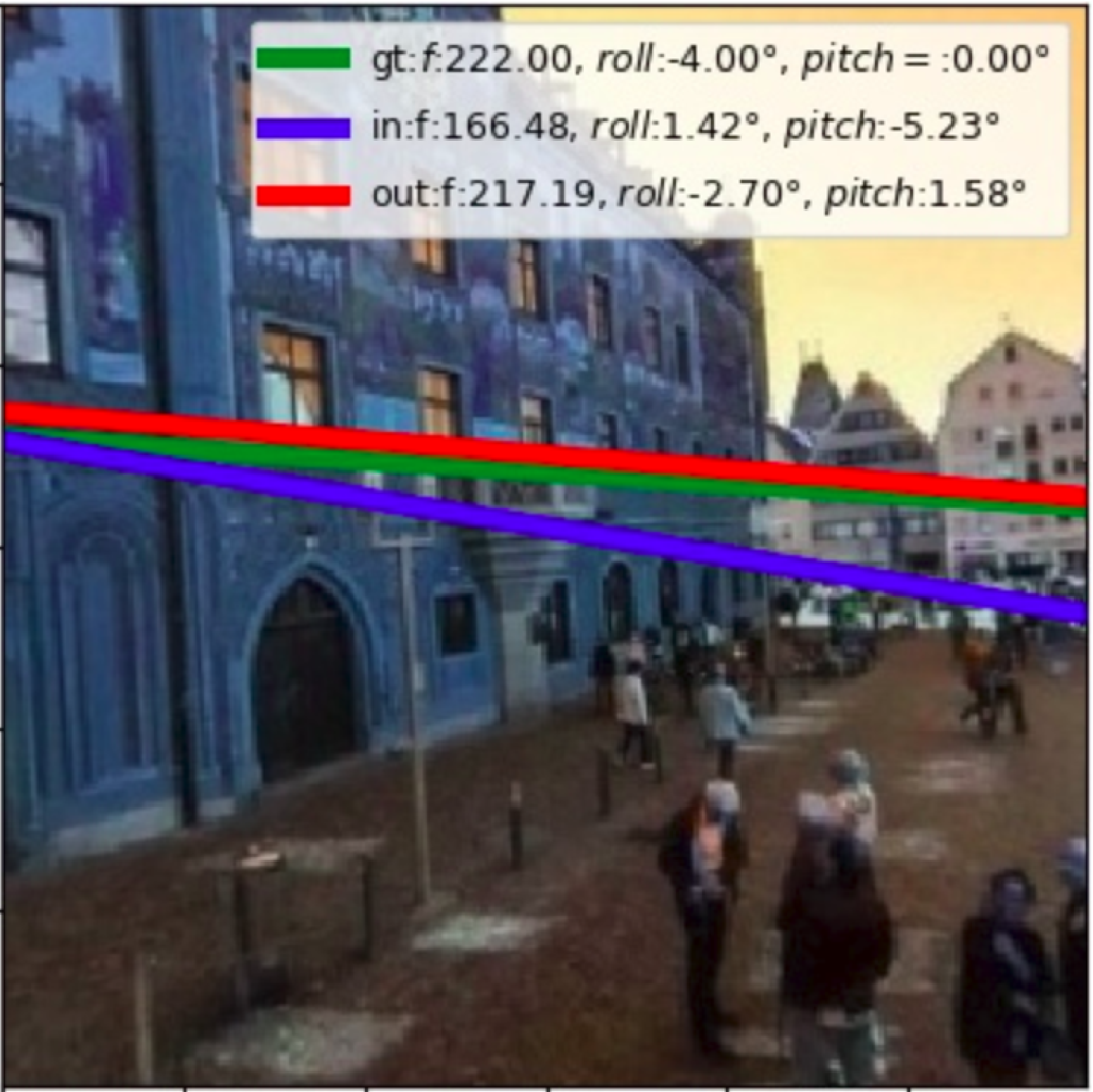}
	\includegraphics[height= 0.32\columnwidth]{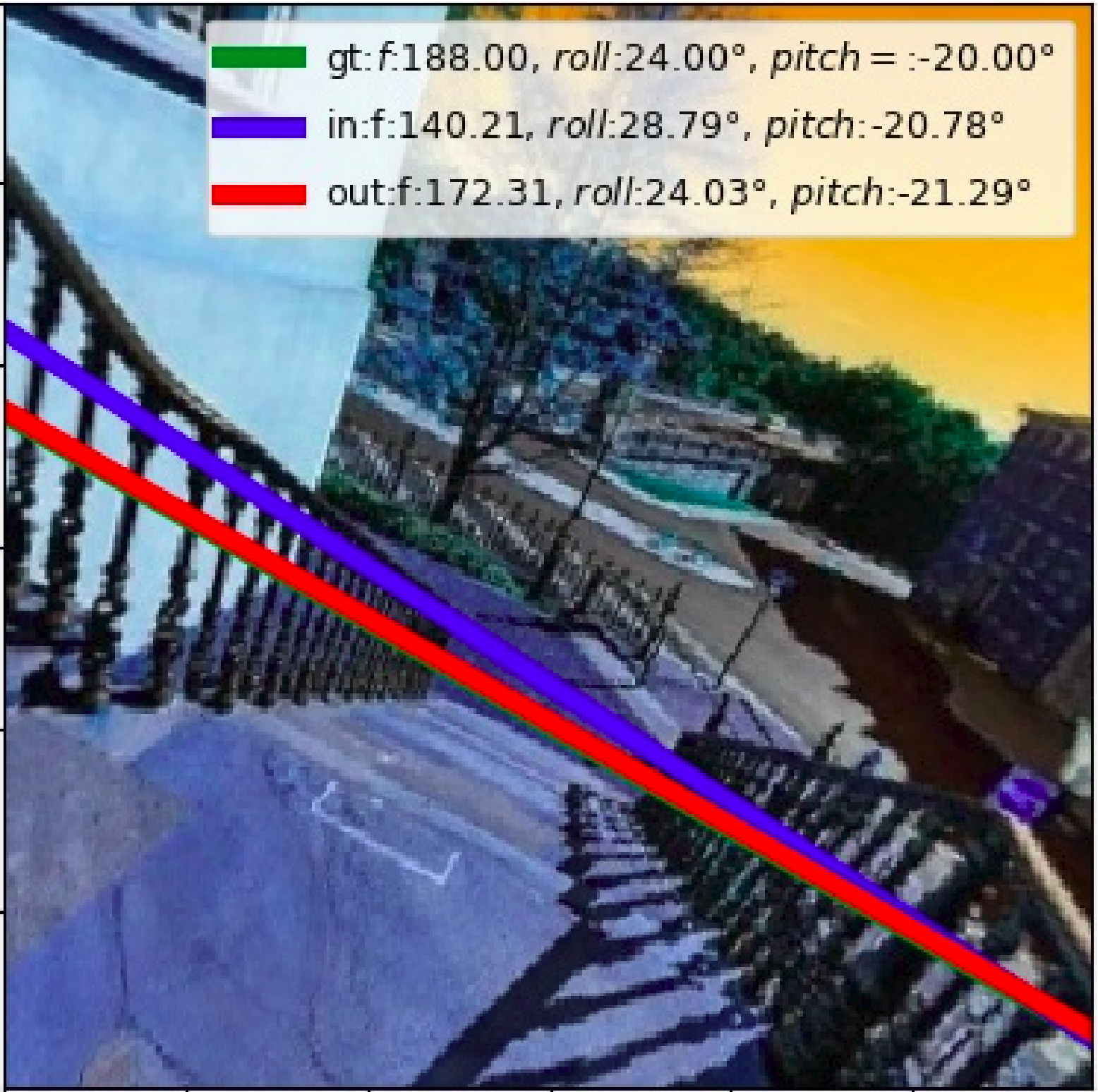}
	\includegraphics[height= 0.32\columnwidth]{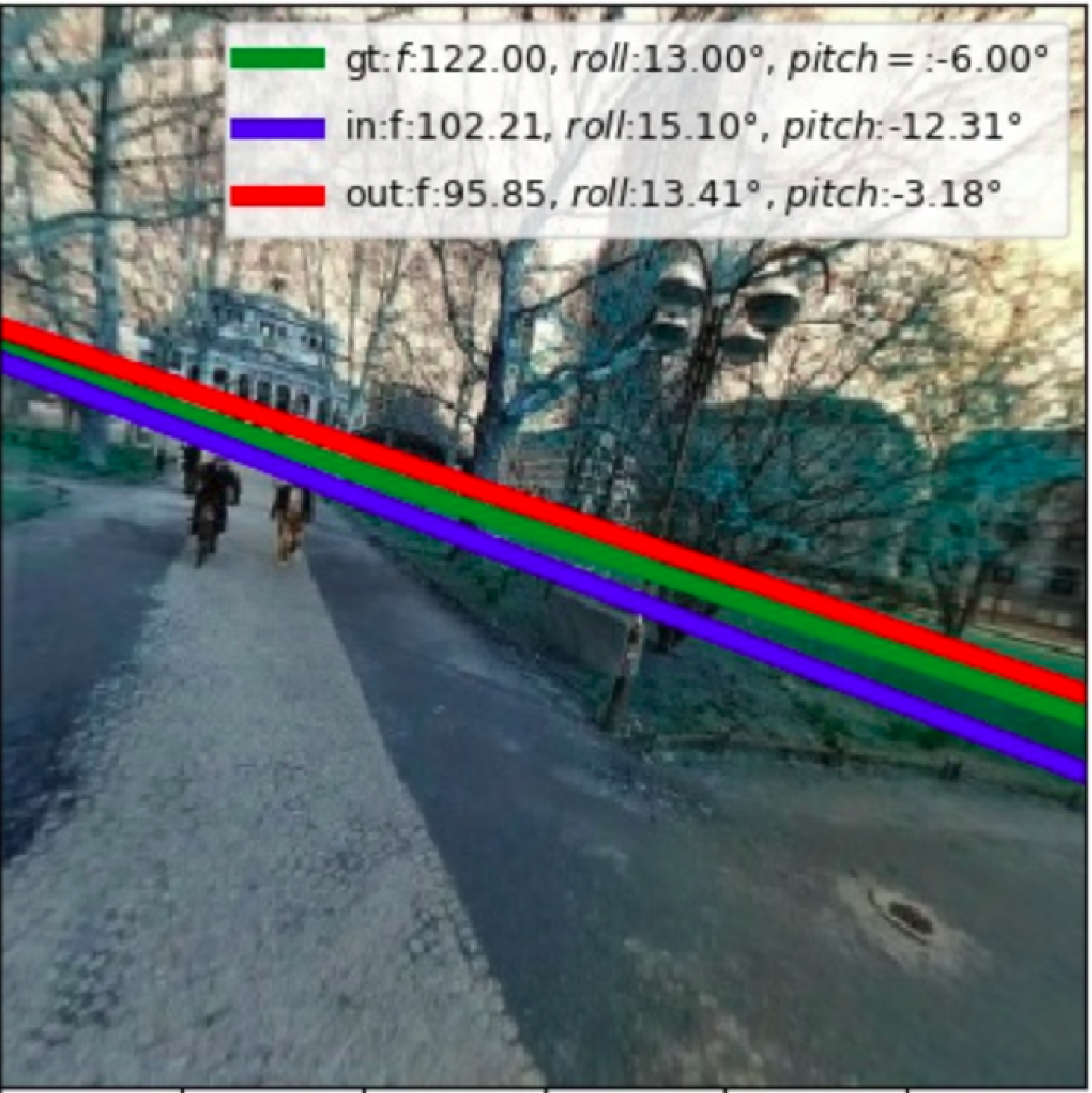}
	\caption{Example of image calibration predictions on indoor (Top row) and outdoor (Bottom row) scene.}
	\label{fig:in_out_lines}
\end{figure}

We note that the two models perform well when tested on the same training data distribution. This result can be observed in Fig.\ref{fig:in_out_lines} where we draw the horizon lines according to the predicted parameters \cite{hold2018perceptual}. We can see that, for indoor images, horizon lines are closer to ground truths than outdoor model lines. Conversely, for outdoor images, we obtain more precise horizon lines with the outdoor model. 

\subsection{Network generalization}
\begin{figure}[h!]
	\centering
	\includegraphics[width= 0.9\columnwidth]{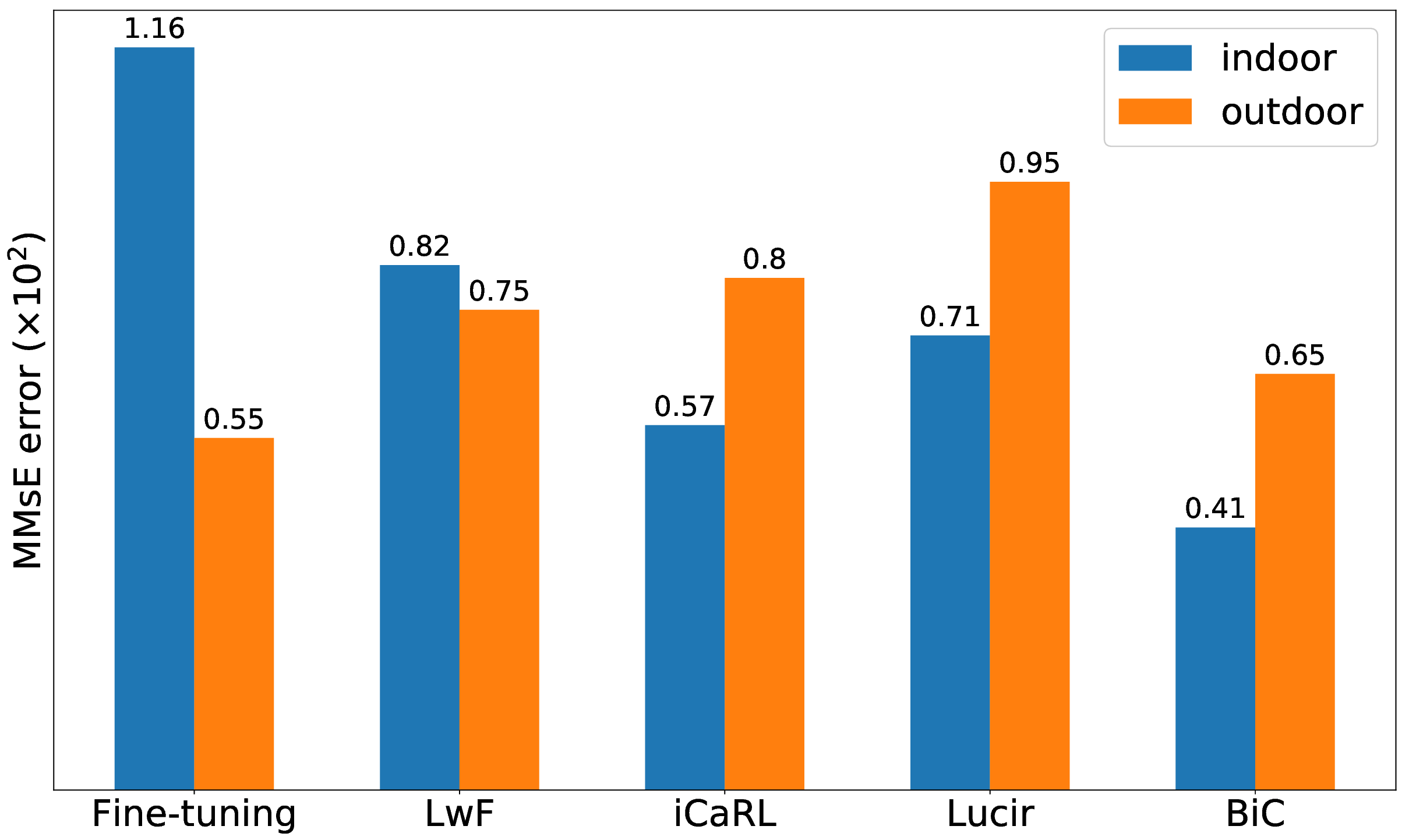}
	\caption{ $\mu MSE$ errors over balanced indoor and outdoor validation sets.} 
	\label{fig:bar}
\end{figure}

The purpose of this work is to estimate the camera parameters from a single image whether for indoor or outdoor scenes. In this section, we compare the performance of several incremental learning techniques (see section.\ref{Incremental_learning}). The aim is to determine which learning model is more suitable to preserve knowledge from different data distributions. Fig. \ref{fig:bar}, compares  $\mu MSE$ errors over indoor and outdoor validation sets.

As shown in the Fig. \ref{fig:bar}, fine-tuning shows a much larger error on indoor data than outdoor data since it learns new data parameters without taking any measures to prevent catastrophic forgetting of indoor data (equivalent to $\lambda =0$ in Eq. \eqref{KD_eq}). In contrast, the indoor error is reduced by $LwF$, which can be seen as a hybrid of fine-tuning and knowledge distillation. Indeed, this technique prevents the deviation of the prediction of the new updated model from the previous one. Further experiments using replay-based methods were performed. These methods store exemplars from previously learned distribution and interleave them with the current learning data. To estimate the number of exemplars, we conducted extensive tests of the $iCaRl$ baseline, by combining the outdoor training set with different percentages of indoor exemplars. Note that, $iCaRL$ with $0\%$ of exemplars is equivalent to $LwF$.
\begin{figure}[ht]
	\centering
	\includegraphics[height= 0.6\columnwidth]{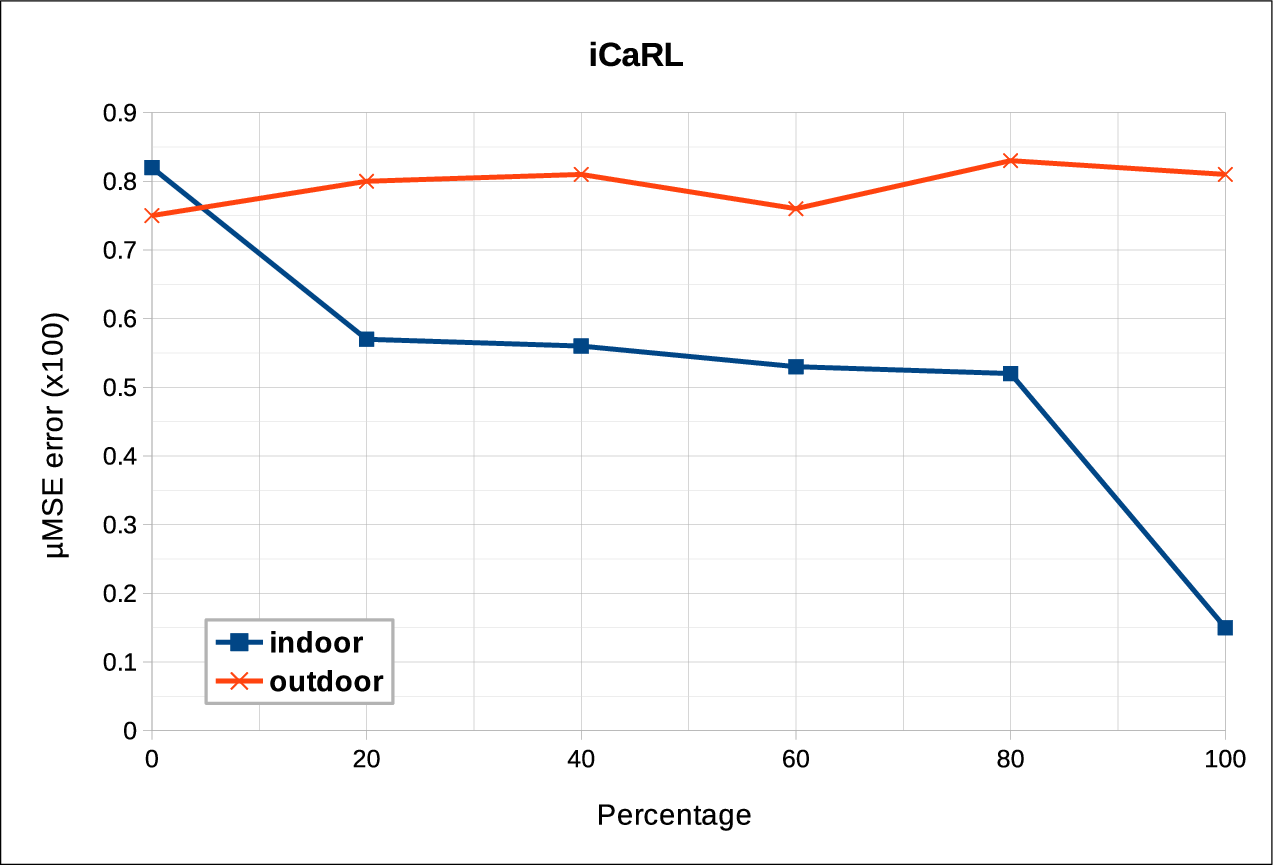}
	\caption{$\mu MSE$ errors over indoor and outdoor using $(\%)$ of exemplars.} 
	\label{fig:bar_icarl}
\end{figure}  

The obtained result from Fig. \ref{fig:bar_icarl} show that $iCaRL$ reduces $LwF$ indoor error from $0.82$ to $0.57$ using $20\%$ of exemplars. Thereafter, the indoor error decreases slightly as the number of exemplars increases. When we kept $60\%$ of exemplars, iCaRL shows a good performance for both indoor and outdoor images. We can explain this by that the indoor images, added during training, contain features shared between indoor and outdoor (edges, contours, etc.). However, we show that uMsE increases when we kept $80\%$ of exemplars, whereas the performance for indoor images keeps increasing as more indoor data are involved during training ($uMsE=0.15$ at $100\%$). The issue can be related to the selected portion of images, which may be largely biased towards indoor images and have no features related to outdoor images.
Therefore, we kept $20\%$ the best trade-off between indoor and outdoor error.

\begin{figure}[ht]
	\centering
	\includegraphics[height= 0.58\columnwidth]{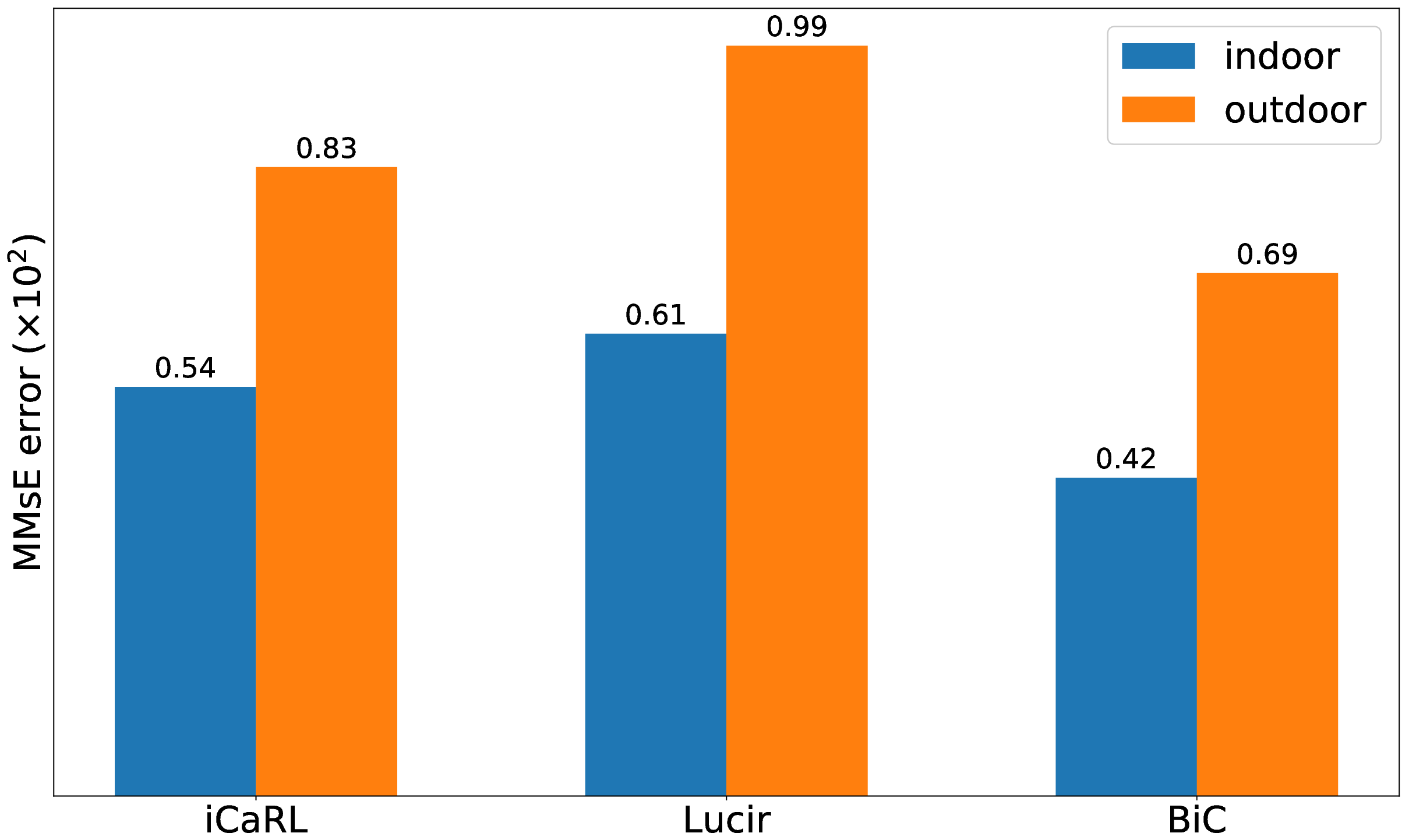}
	\caption{$\mu MSE$ errors over balanced indoor and outdoor test sets ($3584$ images).} 
	\label{fig:bar_test_replay}
\end{figure}

Fig.\ref{fig:bar} compares $iCaR$L with $Lucir$ and $BiC$ methods. All of them utilize knowledge distillation to prevent catastrophic forgetting. As we can see in this figure, $BiC$ performs better than $Lucir$ and $iCaRL$ both on indoor and outdoor scenes. This improvement is achieved using the bias correction layer in the validation stage. In contrast, $Lucir$ degrades the performances of outdoor data ($\mu MSE=0.95$) because it uses a distillation loss on the feature space, which tends to focus on preserving previous knowledge from old data at the expense of learning new inputs. We also note that replay methods perform slightly better on indoor data than outdoor data. This can be justified by the complexity and variety of outdoor images compared to the indoor scenes, which have closer distribution. To confirm our preliminary results, we evaluated the three methods on the new test data (Fig. \ref{fig:bar_test_replay}).

Results demonstrate that the $BiC$ method outperforms $iCaRl$ and $Lucir$, even without using a bias correction layer in the inference. 
\section{Conclusion}
In this study, we address the problem of adapting a single image calibration network to a new distribution while preserving its previous knowledge. Particularly, we adapted the current incremental learning methods $LwF$, $iCarl$, $Lucir$, and $BiC$ to predict camera parameters whether for indoor or outdoor scenes. The comparison results show that $BiC$ performs remarkably well when distilling knowledge, outperforming the evaluated state-of-the-art methods. In future work, we will explore dynamically expandable networks which only change the relevant part of the previously trained network, while still allowing the expansion of its capacity when necessary. 
\bibliographystyle{IEEEbib}
\bibliography{paper}

\begin{thebibliography}{10}

\bibitem{fischer2015image}
F.~Philipp, D.~Alexey, and B.~Thomas,
\newblock ``Image orientation estimation with convolutional networks,''
\newblock in {\em German Conference on Pattern Recognition}. Springer, 2015,
  pp. 368--378.

\bibitem{karsch2014automatic}
K.~Kevin, S.~Kalyan, H.~Sunil, C.~Nathan, J.~Hailin, F.~Rafael, S.~Michael, and
  F.~David,
\newblock ``Automatic scene inference for 3d object compositing,''
\newblock {\em ACM Transactions on Graphics (TOG)}, vol. 33, no. 3, pp. 1--15,
  2014.

\bibitem{mei2007single}
M.~Christopher and R.~Patrick,
\newblock ``Single view point omnidirectional camera calibration from planar
  grids,''
\newblock in {\em Proceedings 2007 IEEE International Conference on Robotics
  and Automation}. IEEE, 2007, pp. 3945--3950.

\bibitem{chen2004camera}
C.~Qian, W.~Haiyuan, and W.~Toshikazu,
\newblock ``Camera calibration with two arbitrary coplanar circles,''
\newblock in {\em European Conference on Computer Vision}. Springer, 2004, pp.
  521--532.

\bibitem{zhu2020single}
Z.~Rui, Y.~Xingyi, H.~Yannick, P.~Federico, E.~Jonathan, S.~Kalyan, and
  C.~Manmohan,
\newblock ``Single view metrology in the wild,''
\newblock in {\em Computer Vision--ECCV 2020: 16th European Conference,
  Glasgow, UK, August 23--28, 2020, Proceedings, Part XI 16}. Springer, 2020,
  pp. 316--333.

\bibitem{workman2015deepfocal}
W.~Scott, G.~Connor, Z.~Menghua, B.~Ryan, and J.~Nathan,
\newblock ``Deepfocal: A method for direct focal length estimation,''
\newblock in {\em 2015 IEEE International Conference on Image Processing
  (ICIP)}. IEEE, 2015, pp. 1369--1373.

\bibitem{bogdan2018deepcalib}
B.~Oleksandr, E.~Viktor, R.~Francois, and B.~Jean-Charles,
\newblock ``Deepcalib: a deep learning approach for automatic intrinsic
  calibration of wide field-of-view cameras,''
\newblock in {\em Proceedings of the 15th ACM SIGGRAPH European Conference on
  Visual Media Production}, 2018, pp. 1--10.

\bibitem{workman2016horizon}
W.~Scott, Z.~Menghua, and J.~Nathan,
\newblock ``Horizon lines in the wild,''
\newblock {\em arXiv preprint arXiv:1604.02129}, 2016.

\bibitem{hold2018perceptual}
H.~Yannick, S.~Kalyan, E.~Jonathan, F.~Matt, G.~Emiliano, H.~Sunil, and
  L.~Jean-Francois,
\newblock ``A perceptual measure for deep single image camera calibration. 2018
  ieee,''
\newblock in {\em CVF Conference on Computer Vision and Pattern Recognition},
  2018, vol.~2, p.~6.

\bibitem{andrew2001multiple}
A.~Alex,
\newblock ``Multiple view geometry in computer vision, by richard hartley and
  andrew zisserman, cambridge university press, cambridge, 2000, xvi+ 607 pp.,
  isbn 0--521--62304--9 (hardback,{\pounds} 60.00).,''
\newblock {\em Robotica}, vol. 19, no. 2, pp. 233--236, 2001.

\bibitem{he2016deep}
H.~Kaiming, Z.~Xiangyu, R.~Shaoqing, and S.~Jian,
\newblock ``Deep residual learning for image recognition,''
\newblock in {\em Proceedings of the IEEE conference on computer vision and
  pattern recognition}, 2016, pp. 770--778.

\bibitem{li2017learning}
L.~Zhizhong and H.~Derek,
\newblock ``Learning without forgetting,''
\newblock {\em IEEE transactions on pattern analysis and machine intelligence},
  vol. 40, no. 12, pp. 2935--2947, 2017.

\bibitem{hou2019learning}
H.~Saihui, P.~Xinyu, L.~C. Change, W.~Zilei, and L.~Dahua,
\newblock ``Learning a unified classifier incrementally via rebalancing,''
\newblock in {\em Proceedings of the IEEE/CVF Conference on Computer Vision and
  Pattern Recognition}, 2019, pp. 831--839.

\bibitem{javed2018revisiting}
J.~Khurram and S.~Faisal,
\newblock ``Revisiting distillation and incremental classifier learning,''
\newblock in {\em Asian conference on computer vision}. Springer, 2018, pp.
  3--17.

\bibitem{Wu_2019_CVPR}
W.~Yue, C.~Yinpeng, W.~Lijuan, Y.~Yuancheng, L.~Zicheng, G.~Yandong, and
  F.~Yun,
\newblock ``Large scale incremental learning,''
\newblock in {\em Proceedings of the IEEE/CVF Conference on Computer Vision and
  Pattern Recognition (CVPR)}, June 2019.

\bibitem{douillard2020podnet}
D.~Arthur, C.~Matthieu, O.~Charles, R.~Thomas, and V.~Eduardo,
\newblock ``Podnet: Pooled outputs distillation for small-tasks incremental
  learning,''
\newblock in {\em Computer Vision--ECCV 2020: 16th European Conference,
  Glasgow, UK, August 23--28, 2020, Proceedings, Part XX 16}. Springer, 2020,
  pp. 86--102.

\bibitem{rebuffi2017icarl}
R.~Sylvestre-Alvise, K.~Alexander, S.~Georg, and L.~Christoph,
\newblock ``icarl: Incremental classifier and representation learning,''
\newblock in {\em Proceedings of the IEEE conference on Computer Vision and
  Pattern Recognition}, 2017, pp. 2001--2010.

\bibitem{shin2017continual}
S.~Hanul, L.~J. Kwon, K.~Jaehong, and K.~Jiwon,
\newblock ``Continual learning with deep generative replay,''
\newblock {\em arXiv preprint arXiv:1705.08690}, 2017.

\bibitem{he2018exemplar}
H.~Chen, W.~Ruiping, S.~Shiguang, and C.~Xilin,
\newblock ``Exemplar-supported generative reproduction for class incremental
  learning.,''
\newblock in {\em BMVC}, 2018, p.~98.

\bibitem{yoon2017lifelong}
Y.~Jaehong, Y.~Eunho, L.~Jeongtae, and H.~Sung Ju,
\newblock ``Lifelong learning with dynamically expandable networks,''
\newblock {\em arXiv preprint arXiv:1708.01547}, 2017.

\bibitem{li2019learn}
L.~Xilai, Z.~Yingbo, W.~Tianfu, S.~Richard, and X.~Caiming,
\newblock ``Learn to grow: A continual structure learning framework for
  overcoming catastrophic forgetting,''
\newblock in {\em International Conference on Machine Learning}. PMLR, 2019,
  pp. 3925--3934.

\bibitem{hinton2015distilling}
H.~Geoffrey, V.~Oriol, and D.~Jeff,
\newblock ``Distilling the knowledge in a neural network,''
\newblock {\em arXiv preprint arXiv:1503.02531}, 2015.

\bibitem{wu2019large}
W.~Yue, C.~Yinpeng, W.~Lijuan, Y.~Yuancheng, L.~Zicheng, G.~Yandong, and
  F.~Yun,
\newblock ``Large scale incremental learning,''
\newblock in {\em International Conference on Computer Vision and Pattern
  Recognition}, 2019, pp. 374--382.

\bibitem{xiao2012recognizing}
X.~Jianxiong, E.~Krista A, O.~Aude, and T.~Antonio,
\newblock ``Recognizing scene viewpoint using panoramic place representation,''
\newblock in {\em 2012 IEEE Conference on Computer Vision and Pattern
  Recognition}. IEEE, 2012, pp. 2695--2702.

\end{thebibliography}

\end{document}